\documentclass[twoside,11pt]{article}

\usepackage[preprint]{jmlr2e}

\usepackage{amsmath, amssymb, bm}
\usepackage{algorithmic, algorithm}
\usepackage{placeins}
\usepackage{booktabs}
\usepackage{xcolor}
\usepackage{fancyvrb}

\DefineVerbatimEnvironment{Code}{Verbatim}{}
\DefineVerbatimEnvironment{CodeInput}{Verbatim}{fontshape=sl}
\DefineVerbatimEnvironment{CodeOutput}{Verbatim}{}
\makeatletter
\newcommand\code{\bgroup\@makeother\_\@makeother\~\@makeother\$\@codex}
\def\@codex#1{{\normalfont\ttfamily\hyphenchar\font=-1 #1}\egroup}
\makeatother
\let\proglang=\textsf
\newcommand{\pkg}[1]{{\fontseries{b}\selectfont #1}}

\hypersetup{
  pdftitle={msPCA: An R Package for Sparse PCA with Multiple Components},
  pdfauthor={Ryan Cory-Wright and Jean Pauphilet},
  pdfkeywords={principal component analysis, sparsity, multiple components, msPCA, Rcpp, R}
}

\usepackage{lastpage}
\jmlrheading{XX}{2026}{1--\pageref{LastPage}}{TBD}{TBD}{TBD}{Ryan Cory-Wright and Jean Pauphilet}

\ShortHeadings{msPCA: Sparse PCA with Multiple Components in R}{Cory-Wright and Pauphilet}
\firstpageno{1}

\begin{document}

\title{\pkg{msPCA}: An \proglang{R} Package for Sparse PCA with Multiple Components}

\author{\name Ryan Cory-Wright \email r.cory-wright@imperial.ac.uk \\
       \addr Analytics, Marketing and Operations\\
       Imperial Business School\\
       \url{https://ryancorywright.github.io}
       \AND
       \name Jean Pauphilet \email jpauphilet@london.edu \\
       \addr Management Science and Operations\\
       London Business School\\
       \url{https://jeanpauphilet.github.io}}


\maketitle

\begin{abstract}
We present \pkg{msPCA}: an open-source \proglang{R} package for sparse principal component analysis with multiple components. It implements an alternating maximization algorithm to generate a set of sparse loading vectors that collectively explain a large fraction of the variance in a dataset, while remaining non-redundant.
The algorithm supports two definitions of non-redundancy: either orthogonality of the loading vectors or zero pairwise correlation between principal components (PCs).
In the reported benchmarks, \pkg{msPCA} solves sparse PCA problems with thousands of features, achieving competitive runtimes while producing sparse components with controlled feasibility violations and a high fraction of variance explained.
\end{abstract}


\section[Introduction]{Introduction}
Principal component analysis \citep[PCA;][]{pearson1901liii,hotelling1933analysis} reduces a $p$-dimensional dataset to $r$ dimensions by projecting onto the leading eigenvectors of the empirical covariance matrix $\bm{\Sigma}$. 
Effectively, each coordinate in the compressed dataset can be expressed as a linear combination of the original variables. 
Beyond interpretability challenges, sample eigenvectors from dense PCA can be inconsistent in high-dimensional regimes when $p/n \to \alpha > 0$ \citep{johnstone2009consistency,deshp2016sparse}, and including every feature in every component complicates downstream analysis \citep{rudin2022interpretable}. Sparse PCA addresses both issues by constraining each loading vector to contain at most $k_t$ nonzero entries. The single-component case is well-studied \citep{d2007direct,yuan2013truncated,bertsimas2022solving}; we focus on the substantially harder $r$-component case, for which guaranteeing non-redundancy across loading vectors is the central challenge. We refer to \citet{bertsimas2022solving} and \citet{cory2022sparse} for comprehensive reviews of the single- and multiple-component cases, respectively.

The package \pkg{msPCA} implements a Lagrangian alternating maximization algorithm for sparse PCA with $r > 1$ components, similar to the one proposed by \citet{cory2022sparse}. The algorithm attaches penalty parameters to the coupling constraints and progressively increases them until feasibility is attained or the iteration budget is reached; at each iteration, the penalized problem decomposes into $r$ independent single-component sparse PCA subproblems, each solved via the Truncated Power Method \citep{yuan2013truncated} against a perturbed covariance matrix that discourages redundancy or overlap between components (see Section~\ref{sec:algorithm}). The algorithm resembles an iterative deflation scheme where a single-component sparse PCA problem on a surrogate matrix is solved at each iteration \citep{mackey2008deflation}. The key difference is that the deflated matrix is induced by an explicit penalty on the non-redundancy constraints, which is progressively increased throughout the algorithm. 
The package is implemented in \proglang{C++} via \pkg{Rcpp} \citep{eddelbuettel2011rcpp} and ships with diagnostic functions for evaluating and displaying solutions.

Compared with \citet{cory2022sparse}, \pkg{msPCA} carries two core software innovations. First, this package supports two types of non-redundancy constraints: \emph{orthogonal loadings} ($\bm{u}_t^\top \bm{u}_{t'} = 0$) and \emph{uncorrelated PCs} ($\bm{u}_t^\top \bm{\Sigma}\,\bm{u}_{t'} = 0$), whereas the algorithm of \citet{cory2022sparse} was originally developed for orthogonality constraints only. Among the existing packages considered here, \pkg{msPCA} is the only package to support both modeling options. Second, the package implements a version of the algorithm that works on the data matrix directly, without the need to compute and store the covariance or correlation matrix, leading to significant computational speed-ups for high-dimensional datasets.
Package \pkg{msPCA} is available from CRAN at \url{https://CRAN.R-project.org/package=msPCA}. A tutorial can be found at \url{https://jeanpauphilet.github.io/msPCA}.

We now compare \pkg{msPCA} with the main packages available for sparse PCA with multiple components. Most of these packages are available in \proglang{R} or \proglang{Python}, with \proglang{R} offering the widest set of active and well-maintained packages for sparse PCA.
These packages cover the main existing methodological approaches:
\pkg{elasticnet} \citep{zou2006sparse} in \proglang{R} and the library \pkg{scikit-learn} \citep{pedregosa2011scikit} in \proglang{Python} implement elastic-net regularization and solve a sequence of single-component problems via a LARS-like algorithm;
the Alternating Manifold Proximal Gradient (A-ManPG) algorithm of \citet{chen2020proximal}, which minimizes a smooth loss with an $\ell_1$ penalty using Riemannian optimization and sequential deflation for multiple components, is implemented in the \proglang{R} package \pkg{amanpg} and the \proglang{Python} package \pkg{sparsepca};
in \proglang{R}, \pkg{PMA} \citep{witten2009penalized} uses a penalized matrix decomposition with a global $\ell_1$ constraint and sequential deflation;
\pkg{sparsepca} \citep{erichson2020sparse} applies a variable-projection framework with an $\ell_1$ penalty;
\pkg{mixOmics} \citep{rohart2017mixomics} implements the sparse PCA approach of \citet{shen2008sparse} via regularized SVD;
\pkg{nsprcomp} \citep{sigg2019nsprcomp} computes sparse PCs via the expectation-maximization algorithm of \citet{sigg2008expectation} combined with deflation or a cumulative formulation.

While \pkg{msPCA} is primarily intended for the \proglang{R} statistical computing ecosystem, sparse PCA is also available in \proglang{Python} via \pkg{scikit-learn} \citep{pedregosa2011scikit}, the dominant scientific machine-learning library for \proglang{Python}. To provide a broad comparison, we include two additional methods beyond the five core \proglang{R} packages.
We compare \pkg{msPCA} against seven competing methods: \pkg{elasticnet} \citep{zou2006sparse}, \pkg{PMA} \citep{witten2009penalized}, \pkg{sparsepca} \citep{erichson2020sparse}, \pkg{mixOmics} \citep{rohart2017mixomics}, and \pkg{nsprcomp} \citep{sigg2019nsprcomp} (all \proglang{R}); \pkg{amanpg} \citep{chen2020proximal} (\proglang{R}, A-ManPG algorithm); and \code{sklearn.decomposition.SparsePCA} from \pkg{scikit-learn} \citep{pedregosa2011scikit} (\proglang{Python}, called from \proglang{R} via the \pkg{reticulate} package \citep{ushey2023reticulate}).
These cover the main methodological approaches: elastic-net regularization, penalized matrix decomposition, variable-projection optimization, regularized SVD, deflation, Riemannian proximal gradient, and dictionary learning.

Regarding sparsity, \pkg{msPCA} allows the user to specify a hard sparsity constraint on each loading vector (of the form $\| \bm{u}_t \|_0 \leq k_t$), which is fairly common across existing packages. Many packages whose algorithms induce sparsity by introducing an $\ell_1$ penalty in the objective, such as \pkg{elasticnet} \citep{zou2006sparse} or \pkg{mixOmics} \citep{shen2008sparse,rohart2017mixomics}, still allow the user to express their requirements in terms of cardinality constraints and automatically calibrate the $\ell_1$ penalty accordingly.
The key distinguishing property of \pkg{msPCA} relates to how overlapping or redundant PCs are discouraged, as summarized in Table~\ref{tab:intro_packages.orth}. First, among the \proglang{R} packages considered here, \pkg{msPCA} is the only package to support both definitions, namely imposing orthogonal loading vectors or uncorrelated PCs. Second, among these packages, \pkg{msPCA} is the only method that provides the user with an explicit control parameter $\eta$ over the violation of these constraints.
Some of the algorithms, such as \pkg{elasticnet} and \pkg{sparsepca} \citep{erichson2020sparse}, internally involve projections onto the set of orthogonal matrices, but these projections are not applied directly to the matrix of loading vectors.
The approaches in \pkg{PMA} \citep{witten2009penalized,witten2009extensions}, \pkg{mixOmics} \citep{rohart2017mixomics}, \pkg{nsprcomp} \citep{sigg2008expectation,sigg2019nsprcomp}, and \pkg{amanpg} apply update (or deflation) rules that guarantee orthogonality in the case of dense PCA but have no guarantees in the sparse PCA case.
In contrast, the alternating maximization algorithm in \pkg{msPCA} assigns an increasing penalty on orthogonality/uncorrelatedness violation, theoretically ensuring asymptotic feasibility when subproblems are solved exactly \citep[see][Section~3]{cory2022sparse}.

\begin{table}[ht]
\centering \footnotesize
\begin{tabular}{llcc}
\hline
Package & Package reference & \multicolumn{2}{c}{Non-redundancy} \\
& & Definition & Control parameter \\
\hline
\pkg{msPCA}      & This paper & orthogonality or uncorrelatedness & tolerance $\eta$  \\
\pkg{elasticnet} & \citet{zou2006sparse}       & orthogonality & None \\
\pkg{scikit-learn}& \citet{pedregosa2011scikit} & orthogonality & None \\
\pkg{PMA}        & \citet{witten2009penalized} & orthogonality & None \\
\pkg{sparsepca}  & \citet{erichson2020sparse}  & orthogonality & None \\
\pkg{mixOmics}   & \citet{rohart2017mixomics}  & orthogonality & None \\
\pkg{nsprcomp}   & \citet{sigg2019nsprcomp} & orthogonality & penalty parameter $\gamma$\\
\pkg{amanpg}     & \citet{chen2020proximal} & orthogonality & None \\
\hline
\end{tabular}
\caption{Main packages for sparse PCA with $r > 1$ components and their implementation of non-redundancy. For sparse PCA, non-redundancy can be expressed as having orthogonal loadings or uncorrelated PCs.}
\label{tab:intro_packages.orth}
\end{table}

The rest of the paper is organized as follows. Section~\ref{sec:background} formalizes the sparse PCA problem with multiple components and introduces the two constraint types supported by \pkg{msPCA}. Section~\ref{sec:algorithm} describes the algorithm, its implementation, and its computational complexity. Section~\ref{sec:package} documents the package functions and illustrates the basic workflow on the \code{mtcars} dataset. Section~\ref{sec:casestudy} presents a case study on S\&P~500 returns comparing both constraint types. Section~\ref{sec:benchmarking} benchmarks \pkg{msPCA} against competing packages on 
real datasets. Section~\ref{sec:conclusion} concludes.

\section[Background]{Sparse PCA with multiple components}\label{sec:background}

\subsection{Problem formulation}

The goal of sparse PCA is to identify $r$ loading vectors $\bm{u}_1, \ldots, \bm{u}_r \in \mathbb{R}^p$ that collectively explain a large share of the variance in the data, while each vector involves only a small number of the $p$ original features. In the single-component case ($r = 1$), this corresponds to solving
\begin{align*}
    \max_{\bm{u} \in \mathbb{R}^p} \quad \bm{u}^\top \bm{\Sigma}\, \bm{u} \quad \text{s.t.} \quad \|\bm{u}\|_2 = 1,\; \|\bm{u}\|_0 \leq k,
\end{align*}
for which many efficient algorithms have been proposed \citep{d2007direct, yuan2013truncated, bertsimas2022solving}. The challenge in the $r$-component case is coordinating the components so that they are non-redundant. In standard (dense) PCA, non-redundancy is ensured by requiring the $r$ leading eigenvectors of $\bm{\Sigma}$ to be mutually orthogonal and their projections to be uncorrelated --- two properties that hold simultaneously for eigenvectors but generally cannot both be enforced in the sparse setting. Package \pkg{msPCA} therefore supports either type of coupling constraint, controlled by the \code{feasibilityConstraintType} argument:

\begin{itemize}
  \item \textbf{Orthogonality} (\code{feasibilityConstraintType = 0}, default): the loading vectors are required to be mutually orthogonal, $\bm{u}_t^\top \bm{u}_{t'} = 0$ for all $t \neq t'$. This is the direct geometric analogue of standard PCA.
  \item \textbf{Zero pairwise correlation} (\code{feasibilityConstraintType = 1}): the projected components are required to be uncorrelated in the data, $\bm{u}_t^\top \bm{\Sigma}\, \bm{u}_{t'} = 0$ for all $t \neq t'$. This ensures that each component captures statistically distinct information.
\end{itemize}

Defining $\bm{C} \in \{\mathbb{I},\, \bm{\Sigma}\}$ to encode the constraint type, the $r$-component problem solved by \pkg{msPCA} takes the unified form
\begin{equation}\label{prob:spca_general}
\begin{aligned}
    \max_{\bm{U} \in \mathbb{R}^{p \times r}} \quad  \sum_{t=1}^r \bm{u}_t^\top \bm{\Sigma}\, \bm{u}_t  \quad
    \text{s.t.} \quad & \bm{u}_t^\top \bm{C}\, \bm{u}_{t'} = 0 \quad \forall\, t \neq t', \\
                      & \|\bm{u}_t\|_2 = 1,\; \|\bm{u}_t\|_0 \leq k_t \quad \forall\, t \in [r],
\end{aligned}    
\end{equation}
where 
$[r] = \{1, \ldots, r\}$. The orthogonality constraint corresponds to $\bm{C} = \mathbb{I}$ and the zero-correlation constraint to $\bm{C} = \bm{\Sigma}$; both reduce to standard PCA when the sparsity constraint is relaxed ($k_t = p$). The two options reflect different modeling priorities. Orthogonality (\code{feasibilityConstraintType = 0}) is the natural choice when the loading vectors will be used as a basis for projection. Zero pairwise correlation (\code{feasibilityConstraintType = 1}) is preferable when the goal is to obtain statistically uncorrelated summaries of the data, for instance for downstream regression or classification. The two options are compared empirically on a real dataset in Section~\ref{sec:casestudy}.

The objective in \eqref{prob:spca_general} corresponds to the sum of per-component variances $\bm{u}_t^\top \bm{\Sigma}\, \bm{u}_t$. Most approaches for sparse PCA with multiple PCs use this objective \citep{zou2006sparse,journee2010generalized,lu2012augmented,vu2013fantope,benidis2016orthogonal,cory2022sparse}. It corresponds to the variance of the orthogonal projection onto the span of $\bm{U}$ only when loading vectors are orthogonal. In general, it corresponds to the sum of the marginal variances of the sparse components.

\subsection{Evaluation metrics}

When assessing the quality of a sparse PCA solution, two quantities are of primary interest. Package \pkg{msPCA} provides dedicated functions for each.

\textbf{Variance explained.} The cumulative fraction of total variance explained (FVE) by a set of loading vectors $\bm{U} = [\bm{u}_1, \ldots, \bm{u}_r]$ is
\[
    \mathrm{FVE}(\bm{U}) = \frac{1}{\mathrm{tr}(\bm{\Sigma})} \displaystyle\sum_{t=1}^r \bm{u}_t^\top \bm{\Sigma}\, \bm{u}_t,
\]
computed by \code{fraction\_variance\_explained()}. Because loading vectors may not be orthogonal, this quantity should be interpreted as a cumulative component-variance score rather than the variance of the orthogonal projection onto the span of $\bm{U}$. The function \code{fraction\_variance\_explained\_perPC()} can be used to return the contribution of each PC, $\bm{u}_t^\top \bm{\Sigma}\, \bm{u}_t / {\mathrm{tr}(\bm{\Sigma})}$, separately. 

\textbf{Feasibility.} The constraint violation measures how far the returned solution is from satisfying the coupling constraints,
\[
    \mathrm{viol}_{\texttt{off}}(\bm{C},\bm{U}) = \sum_{t > t'} \left| \bm{u}_t^\top \bm{C}\, \bm{u}_{t'} \right|,
\]
and is computed by \code{feasibility\_violation\_off()}.

A key shortcoming of many existing sparse PCA packages is that they optimize FVE or sparsity while inadequately enforcing the coupling constraint, so that the returned components are not truly non-redundant. As documented in Section~\ref{sec:benchmarking}, existing sparse PCA packages can return solutions with significant feasibility violations on standard benchmark datasets, although they may return perfectly orthogonal PCs in other cases.

\section[Algorithm]{Algorithm}\label{sec:algorithm}

\subsection{Lagrangian alternating maximization}

The key algorithmic idea is to handle the coupling constraints in Problem~\eqref{prob:spca_general} via a quadratic penalty in the objective, which causes the problem to decompose into $r$ independent single-component sparse PCA subproblems. Specifically, we introduce non-negative penalty parameters $\lambda_{t,t'}$ for each pair $t \neq t'$ (with $\lambda_{t,t'}=\lambda_{t',t}$) and consider the penalized objective
\begin{align}\label{prob:penalized}
    \max_{\bm{U}} \quad \sum_{t=1}^r \bm{u}_t^\top \bm{\Sigma}\, \bm{u}_t - \sum_{t \neq t'} \lambda_{t,t'} \left( \bm{u}_t^\top \bm{C}\, \bm{u}_{t'} \right)^2 \quad \text{s.t.} \quad \|\bm{u}_t\|_2 = 1,\; \|\bm{u}_t\|_0 \leq k_t \quad \forall\, t.
\end{align}
For fixed components $\bm{u}_{t'}$, $t' \neq t$, and fixed penalties $\bm{\lambda}$, the subproblem for $\bm{u}_t$ in \eqref{prob:penalized} reduces to a non-convex single-component sparse PCA problem against the \emph{perturbed} covariance matrix
\begin{align}\label{eq:perturbed}
    \tilde{\bm{\Sigma}}_t = \bm{\Sigma} - \sum_{t' \neq t} \lambda_{t,t'}\, \bm{C}\, \bm{u}_{t'}\bm{u}_{t'}^\top \bm{C}.
\end{align}
Most methods for computing the leading sparse eigenvector of a matrix require the matrix to be positive semidefinite. If the matrix $\tilde{\bm{\Sigma}}_t$ is not positive semidefinite, we can add a diagonal shift $\lambda_0 \mathbb{I}$ to $\tilde{\bm{\Sigma}}_t$, which does not change the optimal solution because all feasible vectors have unit norm. We choose $\lambda_0$ large enough so that $\tilde{\bm{\Sigma}}_t \succeq \bm{0}$, e.g., $\lambda_0 \geq \lambda_{\max}(\bm{C})^2 \sum_{t' \neq t} \lambda_{t,t'}$.

Iterating over $t = 1, \ldots, r$ and progressively increasing the penalties $\bm{\lambda}$ to drive constraint violations toward zero yields the Lagrangian alternating maximization scheme (or iterative deflation scheme) in Algorithm~\ref{alg:main}. Each single-component subproblem is solved via the Truncated Power Method \citep[TPM,][]{yuan2013truncated} described in Algorithm~\ref{alg:TPM}, which alternates between a power step (multiplying by $\tilde{\bm{\Sigma}}_t$) and a truncation step (retaining only the $k_t$ largest-magnitude entries). In practice, TPM often finds high-quality solutions around two orders of magnitude faster than certifiably optimal methods \citep{berk2017, behdin2021sparse}. Our implementation uses the past iterate as the first starting point followed by multiple random restarts, which guard against poor local optima. 

\begin{algorithm}[h]
\caption{Lagrangian Alternating Maximization for Problem~\eqref{prob:spca_general}}
\label{alg:main}
\begin{algorithmic}\normalsize
\REQUIRE Covariance matrix $\bm{\Sigma}$, number of components $r$, sparsity budgets $k_1, \ldots, k_r$, constraint matrix $\bm{C} \in \{\mathbb{I}, \bm{\Sigma}\}$, iterations $L$, feasibility tolerance $\eta$
\STATE Initialize $\bm{u}_t^{(0)} \leftarrow \bm{0}$ for all $t \in [r]$; set $\lambda_{t,t'} \leftarrow 0$ for all $t \neq t'$
\FOR{$\ell = 1, \ldots, L$}
    \FOR{$t = 1, \ldots, r$}
        \STATE Compute $\tilde{\bm{\Sigma}}_t \leftarrow \bm{\Sigma} - \displaystyle\sum_{t' \neq t} \lambda_{t,t'}\, \bm{C}\, \bm{u}_{t'}^{(\ell-1)}\bm{u}_{t'}^{(\ell-1)\top}\bm{C} + \lambda_0\mathbb{I}$
        \STATE Compute $\bm{u}_t^{(\ell)}$ via Algorithm~\ref{alg:TPM} applied to $(\tilde{\bm{\Sigma}}_t,\, k_t)$
    \ENDFOR\vspace{1mm}
    \IF{$\displaystyle \sum_{t} \left| \| \bm{u}_t^{(\ell)} \|^2-1 \right| + \sum_{t > t'} \left|\bm{u}_t^{(\ell)\top} \bm{C}\, \bm{u}_{t'}^{(\ell)}\right| \leq \eta$}
        \STATE Record $\{\bm{u}_t^{(\ell)}\}$ as feasible; update best solution if objective improves
    \ENDIF
    \STATE Increase $\lambda_{t,t'}$ values (see Section~\ref{ssec:implementation})
\ENDFOR
\RETURN best feasible solution found (or least-infeasible solution if none found)
\end{algorithmic}
\end{algorithm}

\begin{algorithm}[h]
\caption{Truncated Power Method with random restarts \citep{yuan2013truncated}}
\label{alg:TPM}
\begin{algorithmic}\normalsize
\REQUIRE Matrix $\tilde{\bm{\Sigma}}$, sparsity budget $k$, iteration limit $L_{TPM}$, time limit $T$
\STATE $\bm{u}_{\mathrm{best}} \leftarrow \bm{0}$
\REPEAT
    \STATE Draw $\bm{u} \sim \mathcal{N}(\bm{0},\mathbb{I})$ 
    \REPEAT
        \STATE $\bm{u} \leftarrow \tilde{\bm{\Sigma}}\bm{u}\, /\, \|\tilde{\bm{\Sigma}}\bm{u}\|_2$ \hfill \textit{(power step)}
        \STATE Zero out all but the $k$ entries of $\bm{u}$ largest in absolute value \hfill \textit{(truncation step)}
        \STATE $\bm{u} \leftarrow \bm{u}\, /\, \|\bm{u}\|_2$
    \UNTIL{$\bm{u}$ converges}
    \IF{$\bm{u}^\top \tilde{\bm{\Sigma}}\, \bm{u} > \bm{u}_{\mathrm{best}}^\top \tilde{\bm{\Sigma}}\, \bm{u}_{\mathrm{best}}$}
        \STATE $\bm{u}_{\mathrm{best}} \leftarrow \bm{u}$; reset iteration count
    \ENDIF
\UNTIL{time limit $T$ exceeded or no improvement after $L_{TPM}$ iterations}
\RETURN $\bm{u}_{\mathrm{best}}$
\end{algorithmic}
\end{algorithm}

\subsection{Implementation details}\label{ssec:implementation}

\textbf{Penalty update.} The penalty parameters $\lambda_{t,t'}$ are initialized to zero and increased progressively across outer iterations, allowing the algorithm to explore freely at first and gradually tighten the feasibility requirement. During the first 15\% of iterations, the increment is proportional to the total constraint violation $\sum_{t > t'} |\bm{u}_t^\top \bm{C}\, \bm{u}_{t'}|$; for the remaining iterations we switch to a ratio-based update proportional to the ratio of the current objective to the current constraint violation, which produces larger and more decisive increases; during the last 25\% of iterations, the step-size coefficient is further increased by a factor of 5 to accelerate final convergence to feasibility. The initial scale of $\lambda_{t,t'}$ is set proportionally to the variance explained by component $t$ at the first iteration, so that all penalty terms are on a comparable scale regardless of the magnitude of $\bm{\Sigma}$. We refer to \citet{cory2022sparse} for a full description and theoretical justification of the update rule.

\textbf{Termination.} Algorithm~\ref{alg:main} stops when (i) the number of outer iterations reaches \code{maxIter} (default: 200), or when (ii) the improvement in objective value between consecutive feasible solutions falls below \code{stallingTolerance} (default: $10^{-8}$). 

\textbf{Feasibility tracking.} At each iteration, Algorithm~\ref{alg:main} checks whether the current solution satisfies the coupling constraint up to the tolerance \code{feasibilityTolerance} (default: $10^{-4}$). The best feasible solution encountered across all iterations is returned. If no feasible solution is found within the iteration budget, the algorithm returns the solution with the smallest observed constraint violation.

\textbf{Software implementation.} All computations are carried out in \proglang{C++} via the \pkg{Rcpp} \citep{eddelbuettel2011rcpp} and \pkg{RcppEigen} \citep{bates2013fast} interfaces, with a lightweight \proglang{R} wrapper providing the user-facing API. To avoid materializing the $p \times p$ perturbed matrix $\tilde{\bm{\Sigma}}_t$ at each inner-loop step, the \proglang{C++} back-end represents it implicitly: each product $\tilde{\bm{\Sigma}}_t \bm{\beta}$ is evaluated as $\bm{\Sigma}\bm{\beta} - \bm{W}(\bm{d} \odot \bm{W}^\top\bm{\beta}) + \lambda_0 \bm{\beta}$, where $\bm{W}$ collects the previously computed components ($\bm{u}_{t'},t' \neq t$ for orthogonal loadings and $\bm{\Sigma}\bm{u}_{t'},t' \neq t$ for uncorrelated PCs) and $\bm{d}$ contains the corresponding scaled penalty coefficients. This eliminates the $O(r p^2)$ matrix-build cost per component update while keeping the per-step cost at $O(p^2)$. When the raw data matrix $\bm{X}$ is provided (\code{type = "X"}), the product $\bm{\Sigma}\bm{\beta}$ is replaced by the two-pass evaluation $\bm{X}^\top(\bm{X}\bm{\beta})/(n-1)$ at cost $O(np)$ instead of $O(p^2)$, which is substantially more scalable when $n \ll p$ and avoids forming the $p \times p$ covariance matrix entirely. After the first outer iteration, the previous iterate $\bm{u}_t^{(\ell-1)}$ serves as a warm start for Algorithm~\ref{alg:TPM}, substantially reducing the number of random restarts required in subsequent iterations. The package is available for Linux, macOS, and Windows.

\subsection{Computational complexity}

The dominant cost of Algorithm~\ref{alg:main} per outer iteration is $r$ calls to Algorithm~\ref{alg:TPM}. With the implicit mat-vec representation described in Section~\ref{ssec:implementation}, applying $\tilde{\bm{\Sigma}}_t$ to a vector costs $O(p^2 + rp)$ when using a covariance or correlation matrix $\bm{\Sigma}$ as an input  (\code{type = "Sigma"}) and $O(np + rp)$ when the raw data matrix $\bm{X}$ is provided (\code{type = "X"}) instead. In both cases, the $O(rp)$ deflation term is negligible for moderate $r$. Each call to Algorithm~\ref{alg:main} performs at most $L_\mathrm{TPM}$ such products, giving a worst-case per-outer-iteration cost of $O( r \min(n,p) p \cdot L_\mathrm{TPM})$. In practice, warm-start initialization from the previous iterate and early convergence detection reduce the effective number of TPM iterations substantially, so the empirical cost is much closer to $O(r \min(n,p) p)$ per outer iteration.

\FloatBarrier

\section[The msPCA Package]{The \pkg{msPCA} package}\label{sec:package}

\subsection{Installation}

Package \pkg{msPCA} is available on the Comprehensive \proglang{R} Archive Network (CRAN) and can be installed and loaded in the usual way:

\begin{Code}
R> install.packages("msPCA")
R> library("msPCA")
\end{Code}

\subsection{Main functions}

\paragraph{\code{mspca()}: Multiple sparse PCA.}
The primary function \code{mspca()} solves Problem~\eqref{prob:spca_general}. Its signature is:

\begin{Code}
R> mspca(M, r, ks, type = c("Sigma", "X"),
+      feasibilityConstraintType = 0,
+      verbose = TRUE, maxIter = 200,
+      feasibilityTolerance = 1e-04, stallingTolerance = 1e-08,
+      timeLimitTPM = 20, maxRestartTPM = 30, minRestartTPM = 20,
+      center = TRUE, scale = TRUE, divisor = c("n-1", "n"),
+      checkPSD = TRUE, symTolerance = 1e-08, psdTolerance = 1e-08)
\end{Code}

The three required arguments are \code{M}, the input matrix; \code{r}, the number of sparse PCs to compute; and \code{ks}, an integer vector of length \code{r} specifying the sparsity budget $k_t$ for each component. The interpretation of \code{M} is controlled by the \code{type} argument: \code{"Sigma"} (the default) treats \code{M} as a $p \times p$ covariance or correlation matrix, while \code{"X"} treats \code{M} as an $n \times p$ raw data matrix. With \code{type = "X"} the algorithm operates directly on the data via the products $\bm{X}^\top(\bm{X}\bm{\beta})$ and never forms the $p \times p$ covariance matrix, which is substantially more scalable when $n \ll p$.

The key optional arguments are \code{feasibilityConstraintType} (\code{0} for orthogonality, \code{1} for zero pairwise correlation), \code{feasibilityTolerance} (the threshold $\eta$), \code{maxIter} (maximum outer iterations), \code{stallingTolerance} (early stopping threshold), and the inner TPM budget controls \code{maxRestartTPM} and \code{minRestartTPM} (random-restart counts for the first and subsequent outer iterations, respectively; see Section~\ref{ssec:implementation}). When \code{type = "X"}, the arguments \code{center}, \code{scale} and \code{divisor} control preprocessing: \code{center = TRUE} (default) mean-centers the columns, \code{scale = TRUE} (default) scales them to unit variance (i.e., the algorithm operates on the correlation matrix), and \code{divisor} selects between the sample covariance (\code{"n-1"}, default) and the population covariance (\code{"n"}).

The function returns seven fields: \code{x\_best} (a $p \times r$ matrix of loading vectors), \code{objective\_value} (the sum of per-component variances $\sum_t \bm{u}_t^\top \bm{\Sigma}\bm{u}_t$), \code{feasibility\_violation} (the constraint violation of the returned solution), \code{runtime} (elapsed time in seconds), \code{variance\_explained} (per-PC explained variance), \code{total\_variance} (trace of the covariance), and \code{inputType} (either \code{"Sigma"} or \code{"X"}). With \code{type = "X"} the result additionally records \code{center}, \code{scale}, \code{divisor}, \code{nObs}, and \code{p}.

\paragraph{\code{tpm()}: Single sparse PC.}
The function \code{tpm()} solves the single-component problem via Algorithm~\ref{alg:TPM} and is useful when only one sparse PC is needed, or as a diagnostic tool to assess the best achievable single-component FVE for a given sparsity budget. Like \code{mspca()}, it accepts either a covariance/correlation matrix or a raw data matrix via the \code{type} argument.

\begin{Code}
R> tpm(M, k, type = c("Sigma", "X"), maxIter = 200, verbose = TRUE,
+    timeLimit = 10, center = TRUE, scale = FALSE, divisor = c("n-1", "n"),
+    checkPSD = TRUE, symTolerance = 1e-08, psdTolerance = 1e-08)
\end{Code}

\subsection{Diagnostic functions}

The package \pkg{msPCA} provides a set of built-in diagnostic functions for evaluating and displaying sparse PCA solutions. These are useful for assessing solution quality and for comparing against solutions from other packages.

\begin{itemize}
  \item \code{fraction\_variance\_explained(C, U)}: combined FVE of the loading matrix \code{U}, as a fraction of $\mathrm{tr}(\bm{\Sigma})$.
  \item \code{fraction\_variance\_explained\_perPC(C, U)}: vector of per-component FVE values.
  \item \code{variance\_explained\_perPC(C, U)}: vector of per-component variance explained (unnormalized).
  \item \code{feasibility\_violation\_off(C, U, feasibilityConstraintType)}: absolute feasibility violation, $\mathrm{viol}_{\texttt{off}}(\bm{C},\bm{U})$ with $\bm{C} = \mathbb{I}$ if \code{feasibilityConstraintType=0} and $\bm{C} = \bm{\Sigma}$ if \code{feasibilityConstraintType=1}.
  \item \code{print(sol, C)}: S3 print method for objects of class \code{"mspca"}. Displays the sparse loading matrix restricted to variables active in at least one component, together with the percentage of variance explained and the number of nonzero loadings per component. When the model was fit with \code{type = "X"}, \code{C} may be omitted and the figures stored inside the result object are used directly.
  \item \code{summary(sol, C, feasibilityConstraintType)}: S3 summary method. Prints a per-PC statistics table (number of nonzero loadings, variance explained, FVE, and cumulative FVE), the upper-triangular pairwise feasibility violation matrix, and the total solver runtime. Like \code{print()}, \code{C} may be omitted for \code{type = "X"} results.
\end{itemize}

\subsection{Worked example}

We illustrate the basic workflow on the \code{mtcars} dataset included in \proglang{R}. The code is available in the replication package. The dataset contains $p = 11$ measurements on 32 car models; we compute the correlation matrix and extract $r = 2$ sparse PCs, each with a sparsity budget of $k = 4$.

\begin{Code}
R> library("msPCA")
R> Sigma <- cor(datasets::mtcars)
R> set.seed(42)
R> res <- mspca(Sigma, r = 2, ks = c(4, 4), verbose = FALSE)
R> print(res, Sigma)
\end{Code}
\begin{CodeOutput}
msPCA solution: 2 sparse PCs
Pct. variance explained: 32.5 28.0
Non-zero loadings per PC:  4 4

Sparse PCs
       [,1]   [,2]
mpg  -0.499  0.000
cyl   0.495  0.000
disp  0.510  0.000
hp    0.000 -0.518
wt    0.495  0.000
qsec  0.000  0.506
vs    0.000  0.494
carb  0.000 -0.482
\end{CodeOutput}

The \code{print()} output omits variables that have zero loadings across all components, making the nonzero loading pattern easy to inspect.

For a richer diagnostic view, we can call \code{summary()}, which additionally reports the pairwise feasibility violation matrix and runtime:

\begin{Code}
R> summary(res, Sigma)
\end{Code}

The cumulative FVE and feasibility violation can also be queried individually:

\begin{Code}
R> feasibility_violation_off(Sigma, res$x_best, 0)
\end{Code}
\begin{CodeOutput}
[1] 0
\end{CodeOutput}
\begin{Code}
R> fraction_variance_explained(Sigma, res$x_best)
\end{Code}
\begin{CodeOutput}
[1] 0.6043866
\end{CodeOutput}

For comparison, we compute the FVE of the first two dense PCs from \code{prcomp()}:

\begin{Code}
R> pca_res <- prcomp(datasets::mtcars, scale. = TRUE)
R> fraction_variance_explained(Sigma, pca_res$rotation[, 1:2])
\end{Code}

\begin{CodeOutput}
[1] 0.8417153
\end{CodeOutput}

The sparse solution captures a smaller fraction of variance than dense PCA (the price of sparsity), but each component loads on only 4 of the 11 variables, making it straightforward to assign physical or mechanical meaning to each PC. 

\subsection{Guidance on parameter choices}

\paragraph{Choosing the sparsity budgets \code{ks}.}
The budgets $k_1, \ldots, k_r$ are the primary tuning parameters. A practical approach is to run \code{mspca()} for a range of values and plot the trade-off between FVE (via \code{fraction\_variance\_explained()}) and sparsity. Domain knowledge often provides a natural guide: if each PC is expected to represent a distinct thematic cluster of features, setting $k_t$ to the anticipated cluster size is a good starting point. 

\paragraph{Choosing the constraint type.}
Orthogonality (\code{feasibilityConstraintType = 0}) is appropriate when the loading vectors are to be used as a projection basis, or when the geometric structure of the components matters. Zero pairwise correlation (\code{feasibilityConstraintType = 1}) is preferable when the primary goal is statistical decorrelation of the projected data. In our experience, the two options yield similar results when $\bm{\Sigma}$ is close to the identity but can differ noticeably for strongly correlated datasets. Section~\ref{sec:casestudy} illustrates both options on financial return data.

\section[Case Study]{Case study: factor analysis of S\&P~500 returns}\label{sec:casestudy}

\subsection{Data and preprocessing}

We illustrate \pkg{msPCA} on daily returns of S\&P~500 constituent stocks, made available under a CC0 1.0 license\footnote{Data available from \url{https://www.kaggle.com/datasets/yash16jr/s-and-p500-daily-update-dataset}}. The code and data for this case study are available in the replication package. The dataset comprises $p = 423$ stocks for which complete price histories are available over the period January~2010 to December~2019, yielding $n = 2{,}515$ trading days. Daily log-returns are computed from adjusted closing prices and stored in the $2{,}515 \times 423$ matrix $\bm{X}$.

S\&P~500 returns are dominated by a ``market factor'' that loads positively on virtually every stock and accounts for a disproportionate share of total variance. To focus on cross-sectional structure (sector and style effects) rather than market-wide movements, we remove the market component before applying \pkg{msPCA}. Specifically, we extract the leading eigenvector $\bm{v}_1$ of the empirical correlation matrix and deflate the correlation matrix as described below:

\begin{Code}
R> library("msPCA")
R> library("RSpectra")
R> library("readr")
R> df_returns <- read_csv("SnP_returns_cleaned.csv") |>
+    dplyr::filter(Date < "2020-01-01")
R> X <- as.matrix(df_returns[, -1])
R> Sigma <- cor(X)
R> v1 <- eigs_sym(Sigma, k = 1, which = "LA")$vectors[, 1]
R> v1 <- v1 / sqrt(sum(v1^2))
R> P  <- diag(length(v1)) - tcrossprod(v1)
R> SigmaR <- crossprod(P, Sigma) %*% P
\end{Code}

\subsection{Sparse factor extraction}

We extract $r = 4$ sparse factors using \pkg{msPCA}, with each factor allowed to load on at most $k$ stocks. We vary $k$ from 5 to 35 in steps of 5 and run the analysis under both constraint types. The following code produces results for the orthogonality constraint (\code{feasibilityConstraintType = 0}); setting \code{feasibilityConstraintType = 1} instead gives the zero pairwise correlation results:

\begin{Code}
R> ks_grid <- seq(5, 35, by = 5)
R> results <- lapply(ks_grid, function(k) {
+    set.seed(42)
+    res <- mspca(SigmaR, r = 4, ks = rep(k, 4), verbose = FALSE,
+                 maxIter = 100, feasibilityConstraintType = 0)
+    data.frame(
+      k = k,
+      fve   = fraction_variance_explained(SigmaR, res$x_best),
+      orth  = feasibility_violation_off(SigmaR, res$x_best, 0),
+      pwcorr = feasibility_violation_off(SigmaR, res$x_best, 1)
+    )
+  })
R> results_df <- do.call(rbind, results)
\end{Code}

\subsection{Comparing constraint types}

\begin{figure}[ht]
  \centering
  \includegraphics[width=0.32\textwidth]{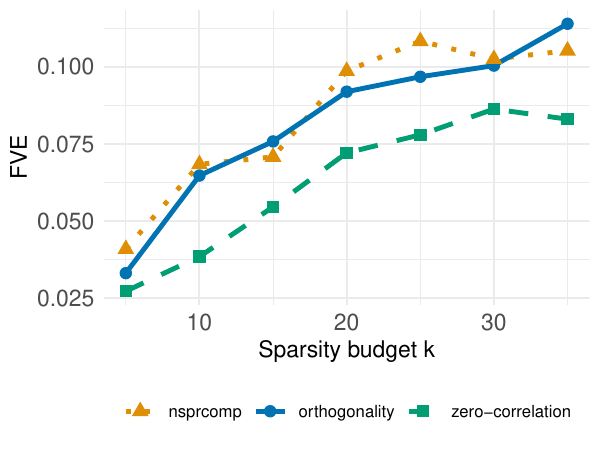}
  \includegraphics[width=0.32\textwidth]{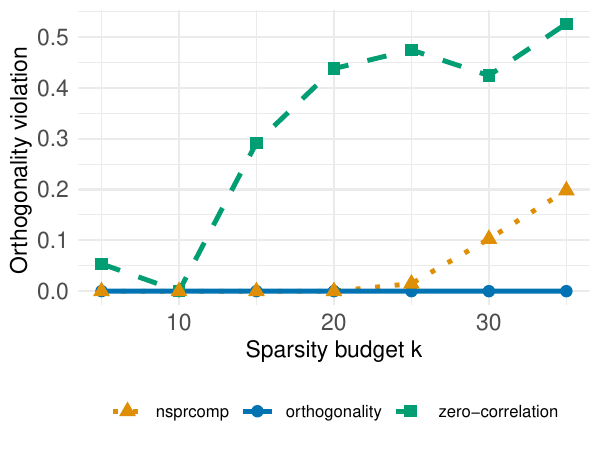}
  \includegraphics[width=0.32\textwidth]{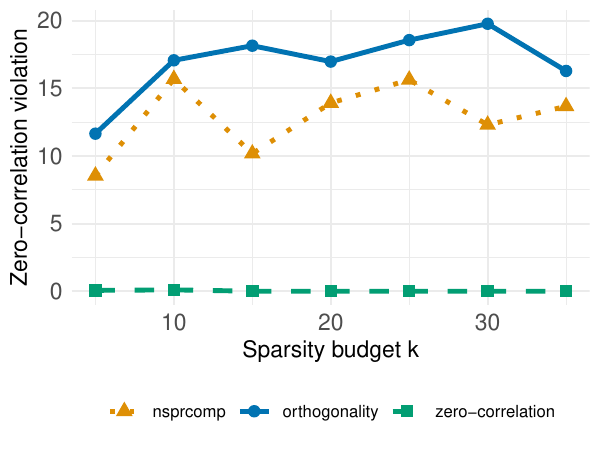}
  \caption{S\&P~500 case study. Left: fraction of variance explained vs.\ sparsity budget $k$.
  Center: orthogonality violation vs.\ $k$.
  Right: uncorrelatedness violation vs.\ $k$. Results are shown for
  \pkg{msPCA} with orthogonality constraints (blue, solid), \pkg{msPCA} with zero pairwise
  correlation constraints (green, dashed), and \code{nsprcomp::nsprcomp} (orange, dotted).
  All methods use $r = 4$ components.}
  \label{fig:snp_results}
\end{figure}

Figure~\ref{fig:snp_results} reports FVE, orthogonality violation ($\mathrm{viol}_{\texttt{off}}(\mathbb{I},\bm{U})$), and uncorrelatedness violation ($\mathrm{viol}_{\texttt{off}}(\bm{\Sigma},\bm{U})$) as a function of $k$, for \pkg{msPCA} with each constraint type. We compare with the function \code{nsprcomp::nsprcomp} of the \pkg{nsprcomp} package \citep{sigg2019nsprcomp} at the same sparsity budgets.

On this dataset, \code{nsprcomp::nsprcomp()} returns orthogonal loading vectors for small-to-moderate sparsity budgets ($k \leq 20$), reflecting the effectiveness of the deflation procedure in producing orthogonal loadings when component supports are sufficiently disjoint. However, as $k$ grows, the orthogonality violation increases, reaching $0.20$ for $k = 35$.

In contrast, \pkg{msPCA} with orthogonality constraints maintains near-zero orthogonality violation across all values of $k$ (at most $10^{-4}$, the default feasibility tolerance), because the penalty on constraint violation is explicitly tightened throughout the algorithm. At the smallest budget ($k = 5$) the violation is $2.2 \times 10^{-4}$, marginally above the default tolerance. In terms of FVE, \code{nsprcomp::nsprcomp()} achieves comparable or slightly higher FVE for most sparsity budgets, with \pkg{msPCA} outperforming it only at the largest budget ($k = 35$).

Neither \code{nsprcomp::nsprcomp()} nor orthogonality-constrained \pkg{msPCA} yields uncorrelated PCs in this case. To obtain uncorrelated PCs, we run \pkg{msPCA} with pairwise correlation constraints instead. Doing so yields PCs with near-zero pairwise correlation, though these are not mutually orthogonal. However, on this dataset, requiring zero pairwise correlation instead of orthogonality is possible only at the expense of a substantially lower FVE.

As illustrated in this case study, these two constraints (orthogonality and zero pairwise correlation) correspond to different feasibility definitions and can lead to meaningfully different factor compositions. \pkg{msPCA} enables the user to explicitly choose and enforce the constraint most relevant to their use case, with predictable behavior across the full range of sparsity levels.

\subsection{Interpretation of sparse PCs}

\begin{figure}[ht]
  \centering
  \includegraphics[width=0.95\textwidth]{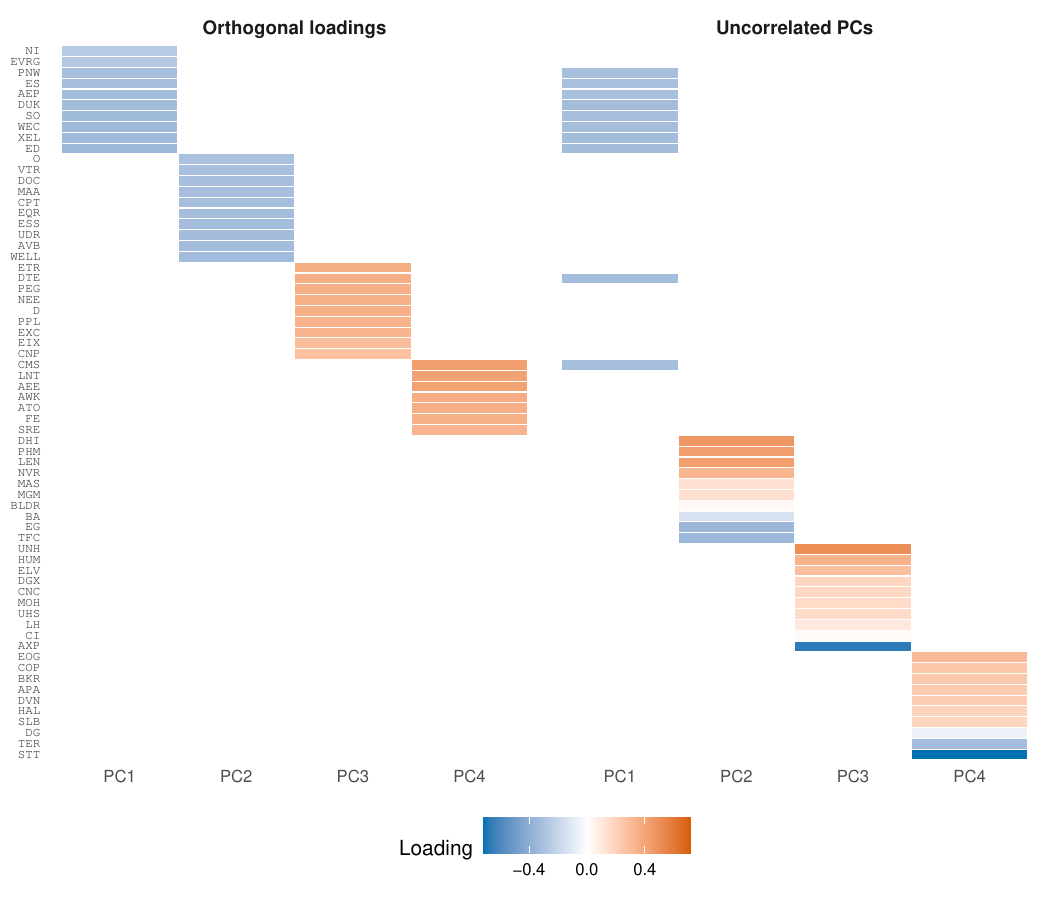}
  \caption{S\&P~500 case study. Loadings of the 4 PCs (sparsity $k=10$) returned by \pkg{msPCA} with orthogonality (left side) or zero-correlation (right side) constraints.}
  \label{fig:snp_loadings}
\end{figure}

We fix $k = 10$ to illustrate the interpretability of sparse PCA and examine the impact of the feasibility constraints used. At this sparsity level each factor loads on 10 stocks out of 423, making it possible to associate each component with an economic theme; sector labels below follow the Global Industry Classification Standard \citep[GICS;][]{msci2023gics}.
Figure~\ref{fig:snp_loadings} shows the nonzero loadings of the 4 PCs returned, depending on the feasibility constraints used.

With orthogonality constraints, the four PCs concentrate entirely within the utility and REIT sectors, with no cross-sector loadings.
PC1 loads on regulated electric utilities (AEP, DUK, ED, ES, EVRG, NI, PNW, SO, WEC, XEL).
PC2 consolidates the REIT segment into a single component spanning residential apartments (AVB, CPT, EQR, ESS, MAA, UDR), healthcare REITs (DOC, WELL, VTR), and net-lease (O).
PC3 captures a second, non-overlapping utility cluster (CNP, D, DTE, EIX, ETR, EXC, NEE, PEG, PPL).
PC4 identifies a third utility subgroup (AEE, ATO, AWK, CMS, FE, LNT, SRE).
Thus the orthogonality constraint fragments the market into three disjoint utility clusters and one diversified REIT basket, with each PC loading exclusively on one sector.

The zero-correlation constraint, on the other hand, finds PCs that are more sector-diverse, each capturing a cross-sector contrast.
PC1 consolidates the entire utility sector into a single broad component (AEP, CMS, DTE, DUK, ED, ES, PNW, SO, WEC, XEL), merging stocks that were split across three separate PCs by the orthogonality constraint.
PC2 contrasts homebuilders and residential construction (DHI, LEN, NVR, PHM, positive loadings; MAS and MGM loading in the same direction) against financials (TFC, EG) and industrials (BA) with negative loadings, capturing the interest-rate sensitivity of the housing sector.
PC3 identifies a managed-care and health-insurance theme: large health insurers and managed-care organizations (UNH, HUM, ELV, MOH, CNC, UHS) load positively alongside diagnostics providers (DGX, LH), while American Express (AXP) carries a large negative loading, reflecting the contrast between healthcare spending and consumer credit.
PC4 loads on oil and gas exploration (APA, COP, DVN, EOG) and oilfield-services companies (BKR, HAL, SLB) positively, against financial custodians (STT) and semiconductor-test equipment (TER) negatively, capturing the energy sector's divergence from interest-rate-sensitive financials and technology.

\section[Benchmarking]{Benchmarking}\label{sec:benchmarking}

We compare \pkg{msPCA} against seven competing packages: \pkg{elasticnet} \citep{zou2006sparse}, \pkg{PMA} \citep{witten2009penalized}, \pkg{sparsepca} \citep{erichson2020sparse}, \pkg{mixOmics} \citep{rohart2017mixomics}, and \pkg{nsprcomp} \citep{sigg2019nsprcomp} (all \proglang{R}); \pkg{amanpg} \citep{chen2020proximal} (\proglang{R}, also available in \proglang{Python} under the name \pkg{sparsepca}); and \code{sklearn.decomposition.SparsePCA} from \pkg{scikit-learn} \citep{pedregosa2011scikit} (\proglang{Python}, called from \proglang{R} via the \pkg{reticulate} package \citep{ushey2023reticulate}).
These cover the main methodological approaches: elastic-net regularization, penalized matrix decomposition, variable-projection optimization, regularized SVD, deflation, Riemannian proximal gradient, and dictionary learning.

Experiments were conducted on a MacBook Air with Apple M2 chip and 24 GB of RAM, with \proglang{R} version 4.6.0 (2026-04-24), \pkg{msPCA} v0.5.0, \pkg{elasticnet} v1.3, \pkg{PMA} v1.2.4, \pkg{sparsepca} v0.1.2, \pkg{mixOmics} v6.36.0, \pkg{nsprcomp} v0.5.1.2, \pkg{amanpg} v0.1.0, \pkg{reticulate} v1.40.0, and \pkg{scikit-learn} v1.6.1 (\proglang{Python} 3.11).

\subsection{Implementation of each method}
Comparing sparse PCA methods is complicated because different packages parameterize sparsity differently. Some methods accept a covariance/correlation matrix $\bm{\Sigma}$, others apply only to the data matrix $\bm{X}$, and some support both. We detail our implementation of each method:

\begin{itemize}
    \item \pkg{msPCA}: We use \code{mspca()} with both \code{type = "Sigma"} and \code{type = "X"} on all four datasets. On the riboflavin dataset ($n \ll p$), the $p \times p$ covariance matrix is rank-deficient, making \code{type = "X"} the natural choice; we nonetheless report results for both modes to quantify the runtime gain.
    \item \pkg{elasticnet}: We use the \code{spca()} function, which takes a list of cardinality budgets (one per PC) as input, and can be applied interchangeably to $\bm{\Sigma}$ or $\bm{X}$. We report results for both input modes on the mtcars, Pitprops, and breast cancer datasets.
    \item \pkg{PMA}: We use the \code{SPC()} function with the \code{orth=TRUE} option, which implements the method of \citet{witten2009penalized}. It takes as inputs the covariance matrix $\bm{\Sigma}$ and a bound on $\sum_{t=1}^{r} \| \bm{u}_t \|_1$, \code{sumabsv}, which we manually tune for each dataset to achieve a per-component sparsity as close as possible to the desired one.
    \item \pkg{sparsepca}: We use the function \code{spca()} applied to $\bm{\Sigma}$, which only allows us to control the magnitude of the $\ell_1$ penalty parameter, \code{alpha}. We manually tune this parameter for each dataset to achieve a total sparsity $\sum_{t=1}^r k_t$ as close as possible to the desired one\footnote{This method tends to return PCs with very unequal sparsity levels, the first PC typically being denser than the subsequent ones.}.
    \item \pkg{mixOmics}: We use the \code{spca()} function, which takes as input the data matrix $\bm{X}$ and the list of sparsity budgets.
    \item \pkg{nsprcomp}: The package provides two functions \code{nsprcomp()} and \code{nscumcomp()}. Both rely on the expectation-maximization algorithm of \citet{sigg2008expectation}. The function \code{nsprcomp()} uses this algorithm to compute the leading sparse PC and then applies the deflation procedure of \citet{mackey2008deflation} to compute the subsequent PCs. It applies to the data matrix $\bm{X}$ and takes as input the list of sparsity levels. The function \code{nscumcomp()} applies expectation-maximization to simultaneously solve for the $r$ PCs. It takes as input the total sparsity budget $\sum_{t =1}^r k_t$ and a penalty parameter $\gamma$ on orthogonality violation (default $\gamma =0$).
    \item \pkg{amanpg}: We use the \code{spca.amanpg()} function applied to the correlation matrix $\bm{\Sigma}$ (\code{type = 1}). The function solves a single-component problem with an $\ell_1$ penalty parameter \code{lambda1} and applies sequential deflation to compute $r$ PCs. We manually tune \code{lambda1} for each dataset to achieve a per-component sparsity as close as possible to the target.
    \item \pkg{scikit-learn}: We use the function \code{sklearn.decomposition.SparsePCA()} called from \proglang{R} via the \pkg{reticulate} package \citep{ushey2023reticulate}, which provides a seamless \proglang{R}/\proglang{Python} interface. The \code{SparsePCA} class computes $r$ components simultaneously using a dictionary-learning approach with an $\ell_1$ penalty \code{alpha} on the loadings. It applies to the data matrix $\bm{X}$. We manually tune \code{alpha} for each dataset to achieve a per-component sparsity close to the target $k$. 
\end{itemize}

For the Pitprops dataset, only the correlation matrix is available, so we generate pseudo-data $\bm{X}$ via \code{MASS::mvrnorm()} with the Pitprops matrix as the population covariance ($n = 500$ observations) and use it as input to the methods that apply to $\bm{X}$ directly. 

For each method, we report the actual number of nonzero loadings for each PC, the FVE (on the original correlation matrix), and the orthogonality violation. We measure (and report) the runtime of each method (excluding hyperparameter tuning) using the \code{system.time()} function. For each method and each dataset, metrics are computed on a single run (with all seeds fixed and provided in the replication scripts).

\subsection{Comparison on real data}\label{sec:realdata}

We benchmark these methods on four datasets.

First, we use the \code{datasets::mtcars} dataset from base \proglang{R} ($p = 11$ variables, 32 observations), which serves as a simple reproducible baseline. We extract $r = 3$ sparse PCs and target a sparsity of $k = 4$ per component. Results are presented in Table~\ref{tab:mtcars}.
\begin{table}[ht]
\centering
\begin{tabular}{l r r r r r}
\hline
Method & $(k_1,k_2,k_3)$ & FVE & Orth.\ violation & Runtime (s) \\
\hline
\code{msPCA::mspca} $(\bm{\Sigma})$  & (4,4,4) & 0.829 & $9.4 \times 10^{-5}$ & 0.323 \\
\code{msPCA::mspca} $(\bm{X})$      & (4,4,4) & 0.829 & $9.4 \times 10^{-5}$ & 0.424 \\
\code{elasticnet::spca} $(\bm{\Sigma})$ & (4,4,4) & 0.753 & 0.117 & 0.165 \\
\code{elasticnet::spca} $(\bm{X})$  & (4,4,4) & 0.754 & 0.117 & 0.189 \\
\code{PMA::SPC}              & (4,4,5) & 0.552 & 0.283 & 0.008 \\
\code{sparsepca::spca}       & (7,4,1) & 0.585 & 0.045 & 0.053 \\
\code{mixOmics::spca}        & (4,4,4) & 0.633 & 0.133 & 0.003 \\
\code{nsprcomp::nsprcomp}    & (4,4,4) & 0.832 & 0.078 & 0.006 \\
\code{nsprcomp::nscumcomp}   & (6,3,3) & 0.772 & 0.245 & 0.033 \\
\code{amanpg::spca.amanpg}  & (7,4,2) & 0.649 & 0.041 & 0.004 \\
\code{sklearn.decomposition.SparsePCA}    & (4,4,3) & 0.827 & $< 10^{-15}$ & 0.028 \\
\hline
PCA                          & (11,11,11) & 0.899 & $< 10^{-15}$ & --- \\
\hline
\end{tabular}
\caption{Comparative performance on the \code{mtcars} dataset ($p = 11$, $r = 3$, $k = 4$).
  \pkg{PMA} uses \code{sumabsv = 1.69} (tuned to achieve 4 nonzeros per component);
  \pkg{sparsepca} uses \code{alpha = 0.004};
  \pkg{amanpg} uses \code{lambda1 = c(10, 3, 0.01)} (tuned per component);
  \code{sklearn.decomposition.SparsePCA} uses \code{alpha = 3.2} (tuned to achieve approximately $k = 4$ nonzeros per component).
  }
\label{tab:mtcars}
\end{table}

Second, we use the Pitprops dataset \citep{jeffers1967two} ($p = 13$ physical measurements on $n=180$ timber specimens), a classic benchmark for sparse PCA methods \citep[e.g., in][]{zou2006sparse}. We extract $r = 6$ sparse PCs and target $k = 4$ nonzero loadings per component as in \citet{zou2006sparse}. We report our results in Table~\ref{tab:pitprops}.
\begin{table}[ht]
\centering
\begin{tabular}{l r r r r r}
\hline
Method & $(k_1, \ldots, k_6)$ & FVE & Orth.\ violation & Runtime (s) \\
\hline
\code{msPCA::mspca} $(\bm{\Sigma})$ & (4,4,4,4,4,4) & 0.786 & $9.7 \times 10^{-5}$ & 1.109 \\
\code{msPCA::mspca} $(\bm{X})$     & (4,4,4,4,4,4) & 0.801 & $9.5 \times 10^{-5}$ & 5.388 \\
\code{elasticnet::spca} $(\bm{\Sigma})$ & (4,4,4,4,4,4) & 0.792 & 0.241 & 0.070 \\
\code{elasticnet::spca} $(\bm{X})$ & (4,4,4,4,4,4) & 0.790 & 0.322 & 0.167 \\
\code{PMA::SPC}            & (4,4,4,7,6,4) & 0.660 & 0.857 & 0.011 \\
\code{sparsepca::spca}     & (8,4,3,1,3,5) & 0.762 & 0.376 & 0.024 \\
\code{mixOmics::spca}      & (4,4,4,4,4,4) & 0.788 & 0.215 & 0.008 \\
\code{nsprcomp::nsprcomp}  & (4,4,4,4,4,4) & 0.800 & 0.626 & 0.023 \\
\code{nsprcomp::nscumcomp} & (5,2,4,5,3,5) & 0.737 & 1.691 & 1.072 \\
\code{amanpg::spca.amanpg} & (8,6,9,2,3,5) & 0.760 & 1.763 & 0.008 \\
\code{sklearn.decomposition.SparsePCA}  & (7,3,4,2,4,4) & 0.895 & 0.733 & 0.043 \\
\hline
PCA        & (13,13,13,13,13,13) & 0.877 & $< 10^{-15}$ & --- \\
\hline
\end{tabular}
\caption{Comparative performance on the \code{Pitprops} dataset \citep{jeffers1967two} ($p = 13$, $r = 6$, target $k = 4$).
  \pkg{PMA} uses \code{sumabsv = 1.727};
  \pkg{sparsepca} uses \code{alpha = 0.004};
  \code{nsprcomp::nscumcomp} uses the default regularization (\code{gamma = 0});
  \pkg{amanpg} uses \code{lambda1 = c(1, 1, 5, 0.4, 0.4, 0)} (tuned per component);
  \code{sklearn.decomposition.SparsePCA} uses \code{alpha = 3.1}.}
\label{tab:pitprops}
\end{table}

Third, we use the breast cancer gene expression data from \citet{chin2006genomic} ($n = 89$ tumor samples, $19{,}672$ genes). Following the preprocessing of \citet{witten2009penalized}, we retain the $p = 500$ genes with the highest marginal variance, yielding a high-dimensional ($p > n$) benchmark for sparse PCA. We set $r = 3$ and target $k = 20$ nonzero loadings per component \citep[a similar sparsity level as in][]{witten2009penalized}. Results are presented in Table~\ref{tab:breast}.
\begin{table}[ht]
\centering
\begin{tabular}{l r r r r}
\hline
Method & $(k_1, k_2, k_3)$ & FVE & Orth.\ violation & Runtime (s) \\
\hline
\code{msPCA::mspca} $(\bm{\Sigma})$ & (20, 20, 20) & 0.093 & $7.5 \times 10^{-5}$ & 27.472 \\
\code{msPCA::mspca} $(\bm{X})$     & (20, 20, 20) & 0.093 & $7.5 \times 10^{-5}$ & 13.382 \\
\code{elasticnet::spca} $(\bm{\Sigma})$ & (20, 20, 20) & 0.043 & 0.040        & 18.130 \\
\code{elasticnet::spca} $(\bm{X})$ & (20, 20, 20) & 0.043 & 0.021            & 0.904  \\
\code{PMA::SPC}            & (14, 17, 27) & 0.028 & 0.001                    & 0.042  \\
\code{sparsepca::spca}     & (46, 13, 8)  & 0.061 & $< 10^{-15}$             & 4.752  \\
\code{mixOmics::spca}      & (20, 20, 20) & 0.077 & $< 10^{-15}$             & 0.031  \\
\code{nsprcomp::nsprcomp}  & (20, 20, 20) & 0.084 & $< 10^{-15}$             & 0.027  \\
\code{nsprcomp::nscumcomp} & (33, 22, 5)  & 0.082 & $< 10^{-15}$             & 0.152  \\
\code{amanpg::spca.amanpg} & (96, 4, 2)   & 0.067 & 0.063                    & 0.252  \\
\code{sklearn.decomposition.SparsePCA}  & (56, 3, 6)   & 0.057 & $< 10^{-15}$             & 0.180  \\
\hline
PCA & (500, 500, 500) & 0.352 & $< 10^{-15}$ & --- \\
\hline
\end{tabular}
\caption{Comparative performance on the breast cancer gene expression dataset \citep{chin2006genomic} ($p = 500$,
  $n = 89$, $r = 3$, target $k = 20$).
  \pkg{PMA} uses \code{sumabsv = 2.92}; \pkg{sparsepca} uses \code{alpha = 0.004} but returns
  unequal sparsity levels across components.
  \code{nsprcomp::nscumcomp} receives a total budget of $k = 60$ and distributes it unevenly
  across components.
  \pkg{amanpg} uses \code{lambda1 = c(8.5, 2.5, 1.5)} (tuned per component);
  \code{sklearn.decomposition.SparsePCA} uses \code{alpha = 11.25}.}
\label{tab:breast}
\end{table}

Finally, we use the riboflavin (vitamin B2) production dataset \citep{buhlmann2014high}, comprising log-transformed expression levels of $p = 4{,}088$ genes in $n = 71$ samples of \textit{Bacillus subtilis} cultures. This is a challenging high-dimensional ($p \gg n$) benchmark: the empirical correlation matrix is rank-deficient (rank $\leq 70$). We set $r = 2$ and target $k = 20$ nonzero loadings per component. Results are presented in Table~\ref{tab:riboflavin}. To avoid time-consuming parameter tuning, we only consider methods that accept exact cardinality budgets.
\begin{table}[ht]
\centering
\begin{tabular}{l r r r r}
\hline
Method & $(k_1, k_2)$ & FVE & Orth.\ violation & Runtime (s) \\
\hline
\code{msPCA::mspca} $(\bm{\Sigma})$ & (20, 20) & 0.008 & $< 10^{-15}$ & 71.986 \\
\code{msPCA::mspca} $(\bm{X})$     & (20, 20) & 0.008 & $< 10^{-15}$ & 1.603  \\
\code{elasticnet::spca} $(\bm{X})$ & (20, 20) & 0.004 & $< 10^{-15}$ & 5.952  \\
\code{mixOmics::spca}               & (20, 20) & 0.006 & $< 10^{-15}$ & 0.241  \\
\code{nsprcomp::nsprcomp}           & (20, 20) & 0.009 & $< 10^{-15}$ & 0.351  \\
\hline
PCA & (4088, 4088) & 0.478 & $< 10^{-15}$ & 0.045 \\
\hline
\end{tabular}
\caption{Comparative performance on the riboflavin dataset \citep{buhlmann2014high} ($p = 4{,}088$,
  $n = 71$, $r = 2$, target $k = 20$). \pkg{msPCA} is run with both \code{type = "Sigma"} and \code{type = "X"}; all other methods use the data matrix directly. We restricted \pkg{msPCA} to \code{maxIter = 100}.}
\label{tab:riboflavin}
\end{table}

\subsection{Discussion}

The results across all four datasets point to several consistent patterns.

First, we find that \code{PMA::SPC()}, \code{sparsepca::spca()}, \code{nsprcomp::nscumcomp()}, \code{amanpg::spca.amanpg()}, and \code{sklearn.decomposition.SparsePCA()} fail to provide satisfactory control over the sparsity of each loading vector, making an apples-to-apples comparison with the other methods difficult.
\code{PMA::SPC()} and \code{sklearn.decomposition.SparsePCA()} control only a global $\ell_1$ bound, while \code{sparsepca::spca()}, \code{nsprcomp::nscumcomp()}, and \code{amanpg::spca.amanpg()} tend to produce unbalanced components with one PC carrying almost all of the sparsity budget.
It is difficult to calibrate these methods to achieve an exact target number of nonzeros per component.
We therefore focus our discussion on \code{msPCA::mspca()}, \code{elasticnet::spca()}, \code{mixOmics::spca()}, and \code{nsprcomp::nsprcomp()}.

Second, among the packages considered here, \code{msPCA::mspca()} is the only method that consistently returns small orthogonality violations on all four datasets, around $10^{-4}$ (the default feasibility tolerance for the method) or below. In contrast, all other packages can sometimes return substantially non-orthogonal components, especially when $p$ is small relative to the total sparsity budget, so trivially orthogonal disjoint-support solutions may not exist. On the \code{Pitprops} data, for example, we observe violations in the $0.2$--$0.6$ range for the other three packages, \pkg{elasticnet}, \pkg{mixOmics}, and \pkg{nsprcomp}. 

In terms of FVE, \code{msPCA::mspca()} and \code{nsprcomp::nsprcomp()} are generally the best-performing methods, followed by \code{mixOmics::spca()} and \code{elasticnet::spca()}, though the relative ranking of these two varies across datasets. Our results suggest that the algorithm in \pkg{msPCA} does not necessarily sacrifice variance explained to achieve orthogonality. 

Finally, our experiments demonstrate that \code{msPCA::mspca()} scales to datasets with $p$ in the thousands in a reasonable amount of time. The \code{type = "X"} interface, which replaces the $O(p^2)$ product $\bm{\Sigma}\bm{\beta}$ with the $O(np)$ two-pass product $\bm{X}^\top(\bm{X}\bm{\beta})/(n-1)$, can yield substantial runtime gains when $n \ll p$, as illustrated on the riboflavin dataset ($p = 4{,}088$, $n = 71$). Using \code{type = "X"} reduces the \pkg{msPCA} runtime from about 72 seconds (\code{type = "Sigma"}) to under 2 seconds (\code{type = "X"}), making it faster than \code{elasticnet::spca()} on this dataset.

\section[Conclusion]{Conclusion}\label{sec:conclusion}

The package \pkg{msPCA} implements and extends an algorithm for sparse PCA with $r > 1$ components proposed in \citet{cory2022sparse}, whose main design feature is that the coupling constraints are explicitly modeled and penalized rather than handled heuristically. The Lagrangian alternating maximization algorithm resembles an iterative deflation procedure where the deflation step depends on increasing penalty parameters that progressively drive constraint violations toward zero across iterations, yielding loading matrices that are simultaneously sparse, high-variance, and designed to be non-redundant. 

A distinguishing feature is the support for two notions of non-redundancy. Orthogonality ($\bm{u}_t^\top \bm{u}_{t'} = 0$) is appropriate when loading vectors serve as a projection basis; zero pairwise correlation ($\bm{u}_t^\top \bm{\Sigma}\,\bm{u}_{t'} = 0$) is preferable when the goal is to obtain statistically uncorrelated scores. The S\&P~500 case study illustrates that the two constraints can yield meaningfully different factor compositions on strongly correlated data, and \pkg{msPCA} allows the user to choose explicitly based on their use case rather than relying on an implicit approximation.

The package has several limitations that point to future work.
First, because of the heuristic nature of the penalty increase scheme, a small fraction of runs may fail to find a feasible solution, though feasibility tracking mitigates this in practice.
Second, the package does not yet support online or streaming settings where observations arrive sequentially.
Future extensions could include robust covariance variants for outlier-contaminated data and a streaming update scheme for sequentially arriving observations.

The package \pkg{msPCA} and its vignette are available at \url{https://CRAN.R-project.org/package=msPCA}.
All scripts and data required to reproduce the examples, figures, and tables are provided in the replication materials.


{\footnotesize
\bibliography{thebib}
}

\end{document}